\documentclass[10pt,twocolumn,letterpaper]{article}

\usepackage[pagenumbers]{iccv}              
\usepackage{caption}
\usepackage{stfloats}
\usepackage{subcaption}
\usepackage{float}
\usepackage{pifont}
\usepackage{multirow}
\usepackage{array}
\usepackage{tikz}
\usetikzlibrary{spy}
\usepackage{url}
\usepackage{colortbl}
\usepackage{threeparttable}
\usepackage{todonotes}
\usepackage{placeins}

\definecolor{iccvblue}{rgb}{0.21,0.49,0.74}
\usepackage[pagebackref,breaklinks,colorlinks,allcolors=iccvblue]{hyperref}

\title{RASMD: RGB And SWIR Multispectral Driving Dataset for Robust Perception in Adverse Conditions}

\author{Youngwan Jin$^{1}$ \quad Michal Kovac$^{2}$ \quad Yagiz Nalcakan$^{1}$ \quad Hyeongjin Ju$^{1}$ \quad\\ Hanbin Song$^{1}$ \quad Sanghyeop Yeo$^{1}$ \quad Shiho Kim$^{1,*}$\\
$^1$ Yonsei University \quad $^2$ Slovak University of Technology\\
$^*$Corresponding author: shiho@yonsei.ac.kr
}

\begin{document}

\twocolumn[{%
\renewcommand\twocolumn[1][]{#1}%
\maketitle
\begin{center}

        \centering
        \begin{tikzpicture}[spy using outlines={circle,red,magnification=2,size=1.1cm, connect spies}]
        \node{\includegraphics[
        width=\textwidth]{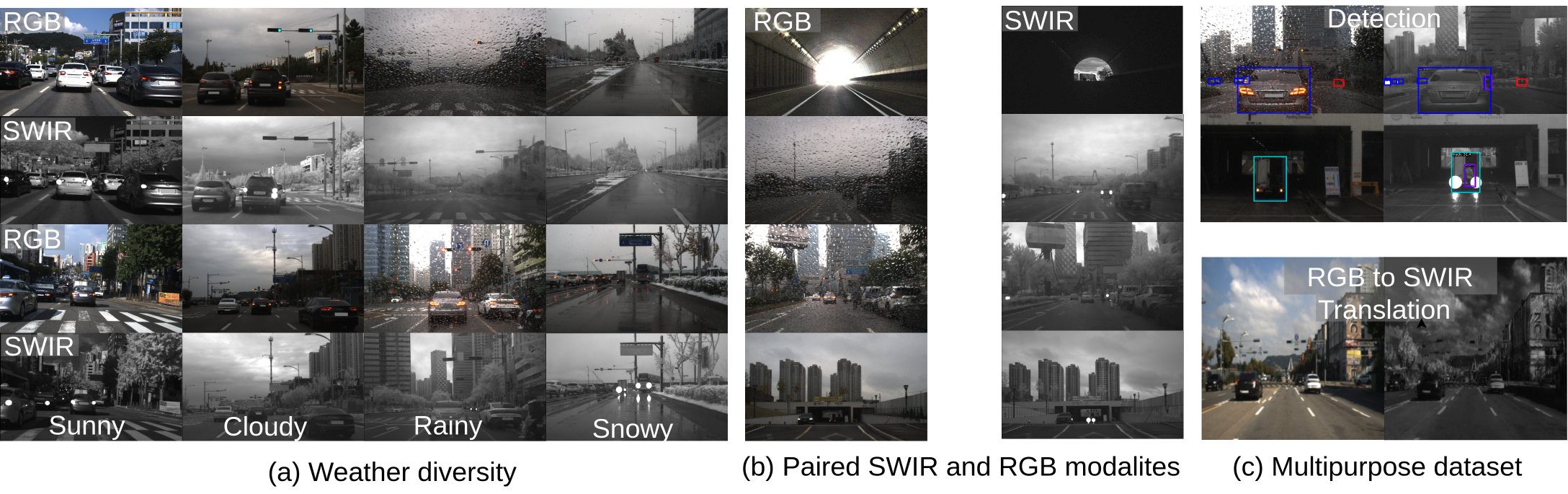}};
        \spy on (0.6,1.9) in node [left] at (2.1,2.2);
        \spy on (3.45,1.9) in node [left] at (3.2,2.2);
        
        \spy on (0.8,0.65) in node [left] at (2.1,1);
        \spy on (3.6,0.65) in node [left] at (3.3,1);
        
        \spy on (0.8,-0.5) in node [left] at (2.1,-0.35);
        \spy on (3.7,-0.5) in node [left] at (3.3,-0.35);
        
        \spy[magnification=7] on (0.55,-1.89) in node [left] at (2.1,-1.7);
        \spy[magnification=7] on (3.42,-1.89) in node [left] at (3.3,-1.7);
        \end{tikzpicture}
 
     \captionof{figure}{RAMSD consists of paired pixel-wise registered SWIR (Short-Wave Infrared) and RGB images captured under various weather conditions \textbf{(a)}. Dual modality provides an advantage in different weather and situations \textbf{(b)}. We provide benchmarks of our dataset for object detection and domain translation tasks \textbf{(c)}.}

     \label{fig:teaser} 

\end{center}%
}] 
\begin{abstract}

Current autonomous driving algorithms heavily rely on the visible spectrum, which is prone to performance degradation in adverse conditions like fog, rain, snow, glare, and high contrast. Although other spectral bands like near-infrared (NIR) and long-wave infrared (LWIR) can enhance vision perception in such situations, they have limitations and lack large-scale datasets and benchmarks. Short-wave infrared (SWIR) imaging offers several advantages over NIR and LWIR. However, no publicly available large-scale datasets currently incorporate SWIR data for autonomous driving. To address this gap, we introduce the RGB and SWIR Multispectral Driving (RASMD) dataset, which comprises 100,000 synchronized and spatially aligned RGB-SWIR image pairs collected across diverse locations, lighting, and weather conditions. In addition, we provide a subset for RGB-SWIR translation and object detection annotations for a subset of challenging traffic scenarios to demonstrate the utility of SWIR imaging through experiments on both object detection and RGB-to-SWIR image translation. Our experiments show that combining RGB and SWIR data in an ensemble framework significantly improves detection accuracy compared to RGB-only approaches, particularly in conditions where visible-spectrum sensors struggle. We anticipate that the RASMD dataset will advance research in multispectral imaging for autonomous driving and robust perception systems. The RASMD dataset is publicly available in \url{https://yonsei-stl.github.io/RASMD/}.

\end{abstract}

\section{Introduction}\label{sec:intro}

In recent advancements towards autonomous driving, the development of a robust and highly accurate vision perception system has become indispensable. Current computer vision algorithms now achieve exceptional accuracy through deep neural networks and data-driven machine learning techniques, leveraging large-scale datasets in training. High-performance systems are benchmarked on massive driving datasets such as Waymo \cite{waymo}, nuScenes \cite{nuscene}, and Open MARS dataset \cite{openmars} alongside large-scale models that ensure the performance.

\begin{figure*}[ht!]
    \centering
    \includegraphics[width=1\linewidth]{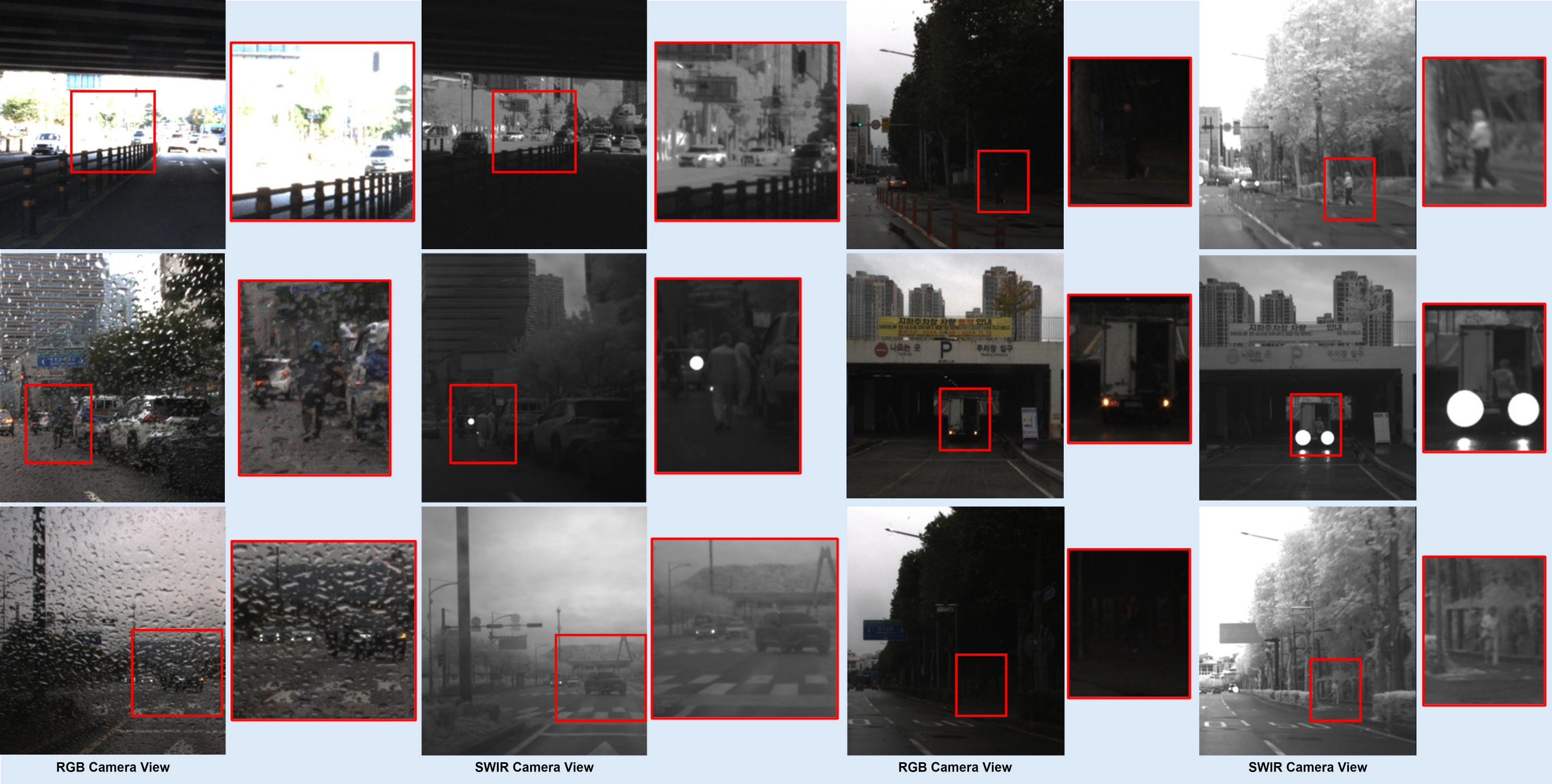}
    \caption{\textbf{Examples from the RASMD dataset:} Each pair shows RGB and SWIR views of the same scene. The SWIR camera demonstrates advantages in challenging conditions, making crucial traffic-related objects visible, which are otherwise difficult or impossible to discern in the RGB images.}
    \label{fig:benefit_of_SWIR}
\end{figure*}

Historically, most of the studied algorithms have relied heavily on the visible spectrum, which is susceptible to performance degradation under adverse conditions like poor weather and low lighting. However, real-world driving environments present a multitude of challenges (Figure \ref{fig:benefit_of_SWIR}) requiring perception systems that remain robust across diverse conditions. To address these limitations, recent research has explored the integration of sensors operating across various spectral domains, such as near-infrared (NIR: 700-1000nm) \cite{nir-detection, IDD_AW, rgb-nir-detection} and long-wave infrared (LWIR: 8000-12000nm) \cite{kaist, kaist_ms2, swir_lwir_detection, swir_detection}. These advancements aim to develop perception systems that maintain reliability and robustness, even under challenging conditions, thereby advancing the frontier of autonomous driving perception capabilities. However, the lack of a well-established large-scale dataset and public benchmark is a challenging problem. The publicly available datasets for autonomous driving are overwhelmingly composed of the visible spectrum band, but they very rarely include imaging beyond the visible spectrum (\Cref{tab:dataset_comparison}).

Incorporating bands beyond the visible spectrum offers certain advantages but also comes with some limitations. For example, LWIR cannot penetrate glass, which limits sensor placement to the only vehicle’s exterior, where exposure to environmental factors complicates maintenance \cite{eswir}. Additionally, LWIR’s low resolution and limited texture contrast hinder detailed scene analysis. Its high sensitivity to temperature further restricts its utility, as it struggles to distinguish between objects with similar thermal properties \cite{lwir_disadvatage, thermal_superres1}. The NIR spectrum demonstrates higher penetration performance compared to the RGB spectrum but still faces scattering challenges in fog, smoke, and haze due to its shorter wavelength range (700–1000 nm) \cite{swir_advantage,swir_advantage1,swir_advantage2}.

In contrast, Short-wave infrared (SWIR: 1000-1700nm) imaging offers several advantages over the limitations seen in NIR and LWIR. Unlike LWIR, SWIR can penetrate glass, allowing it to be installed within the vehicle and protected from environmental exposure. Furthermore, SWIR’s lower sensitivity to temperature variations, coupled with its higher resolution and enhanced texture contrast, allows more detailed scene analysis under diverse environmental conditions than LWIR \cite{eswir}. The longer SWIR wavelengths perform significantly reduced scattering compared to shorter wavelengths like NIR, this enables better penetration through atmospheric challenges such as fog and haze \cite{swir_advantage,swir_advantage1}. These properties make SWIR particularly effective for perception tasks in adverse environments where traditional imaging systems may struggle (\Cref{fig:benefit_of_SWIR}). Despite these advantages, there remains a significant gap in the availability of large-scale SWIR datasets for autonomous driving.
The absence of SWIR data hinders developing and benchmarking algorithms that leverage SWIR’s potential in various driving conditions. To address this gap, we introduce the RASMD dataset, the very first large-scale multispectral dataset that includes paired RGB and SWIR images collected in various locations and diverse weather. Additionally, to validate the effectiveness of our RASMD dataset and SWIR range imaging, we conducted extensive quantitative and qualitative experiments that compared multiple object detection methods and image translation methods.

\noindent Summary of our contribution:
\begin{itemize}
    \item We introduce the RASMD dataset, comprising a total of 100K paired RGB (100K) and SWIR (100K) images, addressing the absence of SWIR datasets for autonomous driving. The data was collected in diverse locations and various weather conditions to support research toward more robust perception systems.  
    \item To validate the utility of the RASMD dataset, we conduct experiments on two downstream tasks: RGB, SWIR object detection, and RGB-SWIR translation. The results demonstrate SWIR’s potential to enhance perception in adverse driving conditions, highlighting our dataset’s value in advancing research on robust vision systems.    
\end{itemize}
\section{Related work}\label{sec:related_work}

\begin{table}[t]
    \scriptsize
    \centering
    \begin{tabular}{lccc}
        \toprule
        \textbf{Dataset} & \textbf{Year} & \textbf{Wavelengths*} & \textbf{$\#$ frames**} \\
        \midrule
        KITTI \cite{kitti} & 2012 & RGB & 15K   \\
        Cityscapes \cite{cityscapes} & 2016 & RGB & 20K \\
        WildDash 2 \cite{wilddash} & 2018 & RGB & 5K   \\
        ApolloScape \cite{apolloscape} & 2019 & RGB & 143K \\
        A2D2 \cite{a2d2} & 2020 & RGB & 392K \\
        A*3D \cite{a3d} & 2020 & RGB & 39K \\
        nuScenes \cite{nuscene} & 2020 & RGB & 1.4M \\
        Waymo Open Dataset \cite{waymo} & 2020 & RGB & 990K \\
        BDD100K \cite{bdd100k} & 2020 & RGB & 100K   \\
        ACDC \cite{acdc} & 2021 & RGB & 3.1K   \\
        Ithaca365 \cite{ithaca365} & 2022 & RGB & 690K \\
        V2V4Real \cite{v2v4real} & 2023 & RGB & 40K \\
        Zenseact Open Dataset \cite{zenseact} & 2023 & RGB & 100K \\
        Open MARS Dataset \cite{openmars} & 2024 & RGB & 1.4M \\
        \midrule
        KAIST \cite{kaist} & 2015 & RGB, LWIR & 95K \\
        CVC-14 \cite{CVC14} & 2016 & RGB, LWIR & 7.7K \\
        RANUS \cite{ranus} & 2018 & RGB, NIR & 40K \\
        LLVIP \cite{LLVIP} & 2021 & RGB, LWIR & 15K \\
        MFnet \cite{mfnet} & 2021 & RGB, LWIR & 1.5K \\
        FLIR \cite{flir} & 2022 & RGB, LWIR & 10K \\
        MS2 \cite{kaist_ms2} & 2022 & RGB, NIR, LWIR & 195K \\
        FMB \cite{FMB} & 2023 & RGB, LWIR & 1.5K \\
        IDDAW \cite{IDD_AW} & 2024 & RGB, NIR & 5K \\
        InfraParis \cite{infraparis} & 2024 & RGB, LWIR & 7.3K \\
        \midrule
        \rowcolor{gray!20}
        \textbf{Ours} & 2024 & RGB, \textbf{SWIR} & 100k \\
        \bottomrule
    \end{tabular}
   \caption{Comparison of datasets used for autonomous driving tasks. The "RGB" definition is used to indicate visible range imaging. The total frame amount is given for each dataset.}
   \label{tab:dataset_comparison}
\end{table}

\subsection{Autonomous Driving Datasets}

\noindent \textbf{Visible Spectrum Datasets:} Autonomous driving systems have long relied on datasets captured in the visible spectrum, as these RGB datasets form the cornerstone for training and validating perception algorithms in tasks such as object detection, segmentation, and scene understanding. Early datasets like KITTI \cite{kitti} and Cityscapes \cite{cityscapes} have been instrumental, with KITTI's 15K frames covering diverse tasks and Cityscapes’ 20K frames offering dense semantic segmentation in urban environments. Over time, larger and more varied datasets have emerged, like BDD100K \cite{bdd100k} with 100K frames, which encompasses a wide range of geographic locations, annotations for various tasks, times of day, and weather conditions, supporting more generalized model training.

Recent datasets such as Waymo Open Dataset \cite{waymo} and nuScenes \cite{nuscene} both surpass a million frames, providing extensive sensory data, including lidar and radar, alongside RGB. Datasets like ApolloScape \cite{apolloscape} with 143K frames and A2D2 \cite{a2d2} with 392K frames focus on dense urban traffic scenarios, supporting tasks from object detection to lane marking. For adverse weather conditions, ACDC \cite{acdc} provides 3.1K frames specifically curated to evaluate model robustness in fog, rain, and low-light scenarios. Datasets such as Ithaca365 \cite{ithaca365}, V2V4Real \cite{v2v4real}, and the Zenseact Open Dataset \cite{zenseact} continue to expand the range of real-world conditions represented, including seasonal changes and challenging environments. 

\noindent \textbf{Infrared Imaging Datasets:} While visible spectrum datasets provide a strong foundation for autonomous driving research, several datasets incorporating NIR and LWIR imaging for autonomous driving and computer vision tasks have been published by aiming to overcome the limitations of visible spectrum imaging in adverse conditions.  
For autonomous driving tasks, the RANUS \cite{ranus} dataset offers synchronized RGB and NIR data captured in diverse urban settings for the benchmarking of multi-modal semantic segmentation methods to support the development of algorithms that leverage near-infrared information for more reliable segmentation and under varying light conditions. The IDDAW dataset \cite{IDD_AW} also includes NIR data with a focus on adverse weather scenarios, for exploring detection and semantic segmentation under challenging environmental conditions.
Several autonomous driving datasets have incorporated LWIR data to assess the effectiveness of thermal range for robust perception models under varied environmental conditions. The KAIST Multispectral Pedestrian Detection Benchmark \cite{kaist}, CVC-14 \cite{CVC14} and LLVIP \cite{LLVIP} were early and influential datasets in this area, combining RGB and LWIR modalities to address pedestrian detection challenges, especially in low light conditions. MFNet \cite{mfnet} and FLIR ADAS \cite{flir} datasets expanded on this work by offering synchronized RGB-LWIR frames specifically designed for automotive applications, with again an emphasis on pedestrian and vehicle detection in low-light settings. FMB \cite{FMB} and InfraParis \cite{infraparis} are more recent additions that provide diverse environmental contexts to support the development of multispectral perception systems that combine thermal imaging with visible-spectrum data.

Despite the increasing availability of NIR and LWIR datasets, there is a notable absence of public datasets incorporating SWIR data. To address this gap, we developed the RASMD dataset, the first large-scale, synchronized RGB-SWIR dataset. RASMD is intended to complement existing NIR and LWIR resources by providing unique SWIR imagery that enhances robust perception in conditions like rain, snow, and low and backlight situations. 

\subsection{Image to image translation}

Image-to-image (I2I) translation is a key computer vision task focused on converting images from one domain to another while preserving structural and content details \cite{cyclegan,pix2pix,stargan,attention_i2i}. Initial approaches like Pix2pix \cite{pix2pix} relied on paired datasets to learn mappings with conditional GANs, while Pix2pixHD \cite{pix2pixhd} introduced high-resolution synthesis using multi-scale discriminators. CycleGAN \cite{cyclegan} introduced cycle consistency loss for unpaired data, enabling translation without paired datasets. Recently, BBDM \cite{bbdm} used diffusion processes to enhance translation stability and diversity, addressing limitations like mode collapse in GANs.

In infrared (IR) imaging, a primary challenge is the scarcity of labeled datasets, which has led to approaches for generating synthetic IR images from RGB data. For instance, Pix2pix has been adapted for RGB-to-NIR translation in agriculture \cite{rgb2nir}, while C2SAL \cite{c2sal} applies style transfer for NIR generation in driving scenes. Models like ThermalGAN \cite{thermalgan} and InfraGAN \cite{infragan} generate synthetic LWIR images for thermal IR. Despite these advancements, a gap remains in RGB-SWIR paired datasets, limiting progress in SWIR-specific applications. We evaluated IR range image translation methods with our spatially aligned RGB-SWIR images. 

\subsection{Multi-modal Object Detection}

Object detection is essential in autonomous driving, where identifying road users, interpreting traffic signs, and avoiding obstacles is critical. Conventional vision-based methods, such as Faster R-CNN \cite{faster_rcnn}, SSD \cite{ssd} and Transformer-based approaches like DETR \cite{detr} and its variants \cite{dino, deformable_detr, conditional_detr, codetr}, have been widely adopted for these tasks. However, detection accuracy tends to decline in adverse conditions (e.g., fog, rain, low light) when relying solely on the visible spectrum.

Several studies have explored multispectral imaging for object detection, aiming to address limitations of the visible spectrum \cite{nir-detection, swir_detection, swir_detection2, swir_lwir_detection}. Yu \textit{et al.} \cite{swir_detection2} introduced a three-channel SWIR imaging system with a liquid crystal tunable filter (LCTF) to enhance object detection in hazy conditions. This system uses the YOLOv3 model combined with an RL (recognition and localization) score to select optimal SWIR bands for recognizing objects. Pavlović \textit{et al.} \cite{swir_detection} developed a long-range SWIR-based surveillance setup for foggy environments, using cross-spectral annotation to automatically label SWIR images by transferring visible detections within a multi-sensor configuration. Govardhan and Pati \cite{nir-detection} created a nighttime pedestrian detection system using NIR images, combining Haar Cascade and HOG-SVM classifiers to reduce false positives.

Ensemble and fusion techniques for RGB-multispectral detection also show promise. Li \textit{et al.} \cite{li_multimodal_detection_2023} proposed a confidence-aware framework (CMPD) that combines RGB and thermal data for pedestrian detection, applying Dempster’s rule for data fusion. Karasawa \textit{et al.} \cite{multispectral} achieved a 13\% mAP improvement by incorporating RGB and multiple infrared bands. Similarly, Chen’s ProbEn \cite{chen_multimodal_detection_2022} framework, which ensembles RGB and thermal detection streams, demonstrated significant performance gains on KAIST and FLIR benchmarks.


\section{RASMD (RGB And SWIR Multispectral Driving Dataset)}
\begin{figure*}[ht]
    \centering
    \begin{minipage}{0.3\textwidth}
        \centering
        \includegraphics[width=\linewidth]{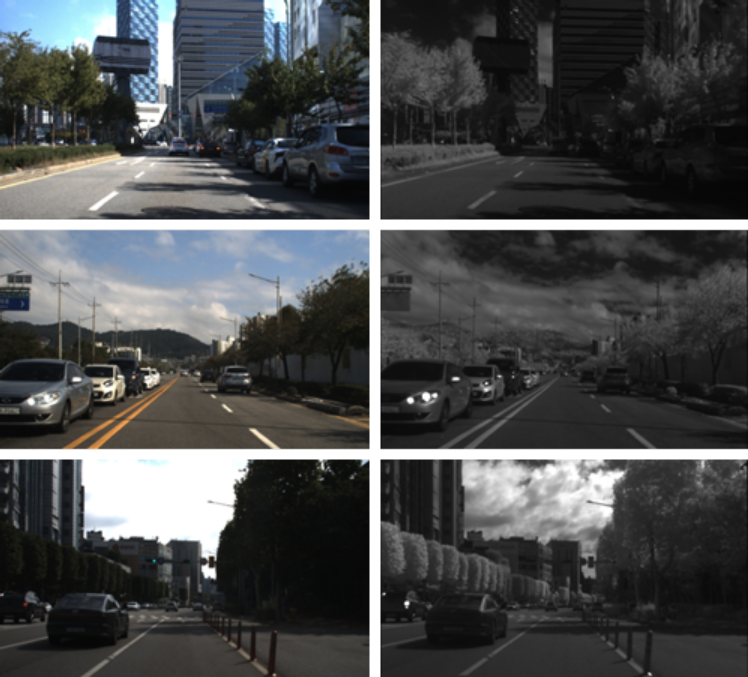}
        \subcaption{Driving Scene "Urban"}
    \end{minipage}
    \begin{minipage}{0.3\textwidth}
        \centering
        \includegraphics[width=\linewidth]{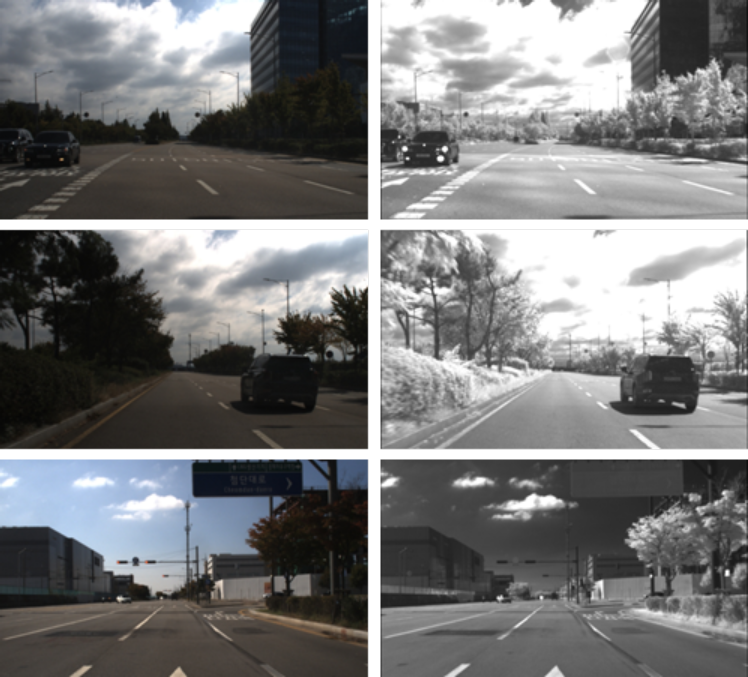}
        \subcaption{Driving Scene "Suburban"}
    \end{minipage}
    \begin{minipage}{0.3\textwidth}
        \centering
        \includegraphics[width=\linewidth]{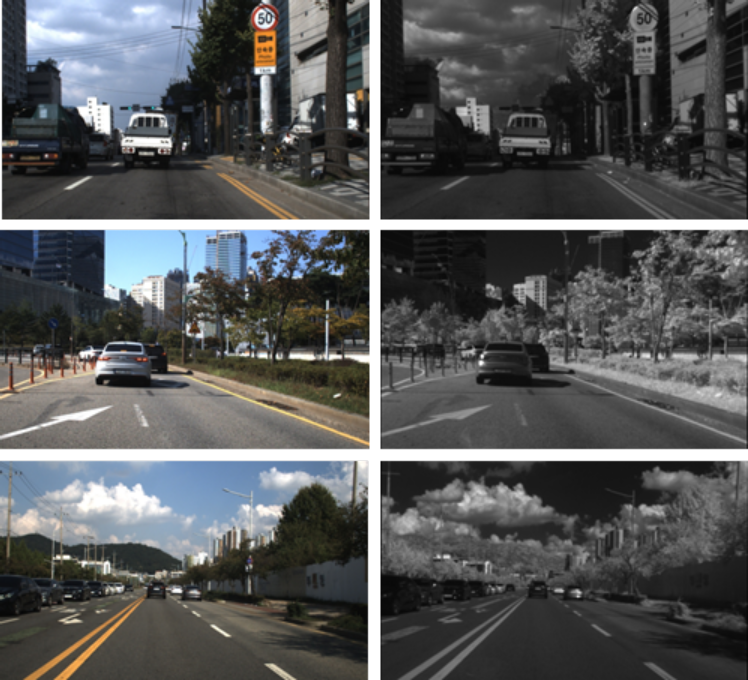}
        \subcaption{Driving Scene "Sunny"}
    \end{minipage}

    \vspace{0.2cm}

    \begin{minipage}{0.3\textwidth}
        \centering
        \includegraphics[width=\linewidth]{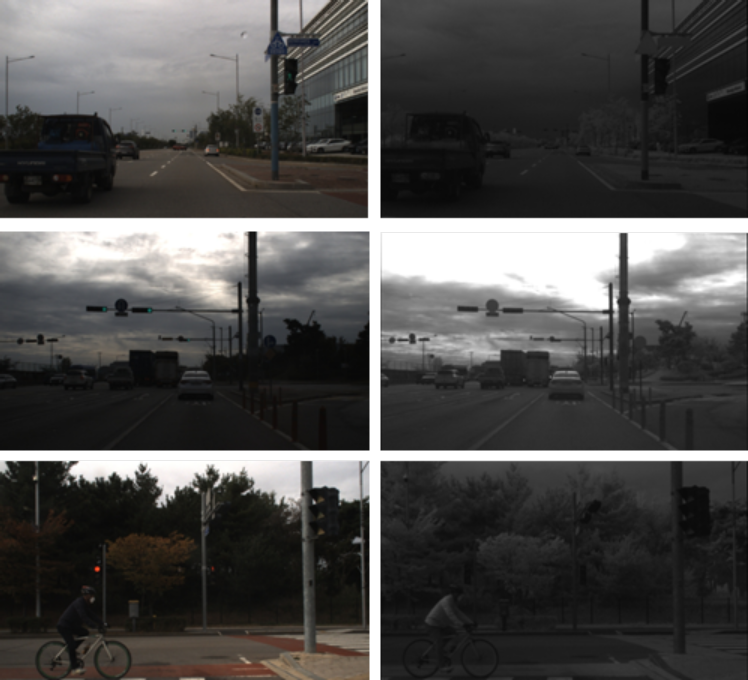}
        \subcaption{Driving Scene "Cloudy"}
    \end{minipage}
    \begin{minipage}{0.3\textwidth}
        \centering
        \includegraphics[width=\linewidth]{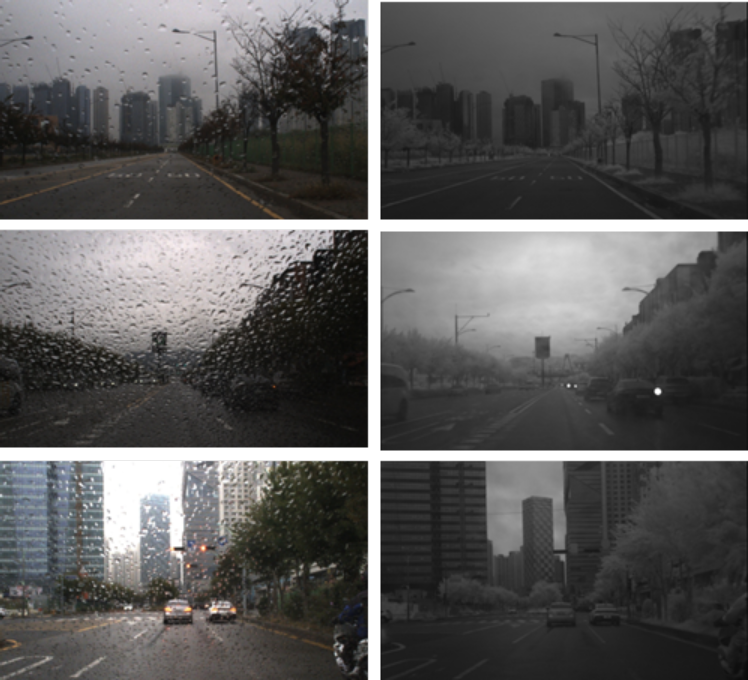}
        \subcaption{Driving Scene "Rainy"}
    \end{minipage}
    \begin{minipage}{0.3\textwidth}
        \centering
        \includegraphics[width=\linewidth]{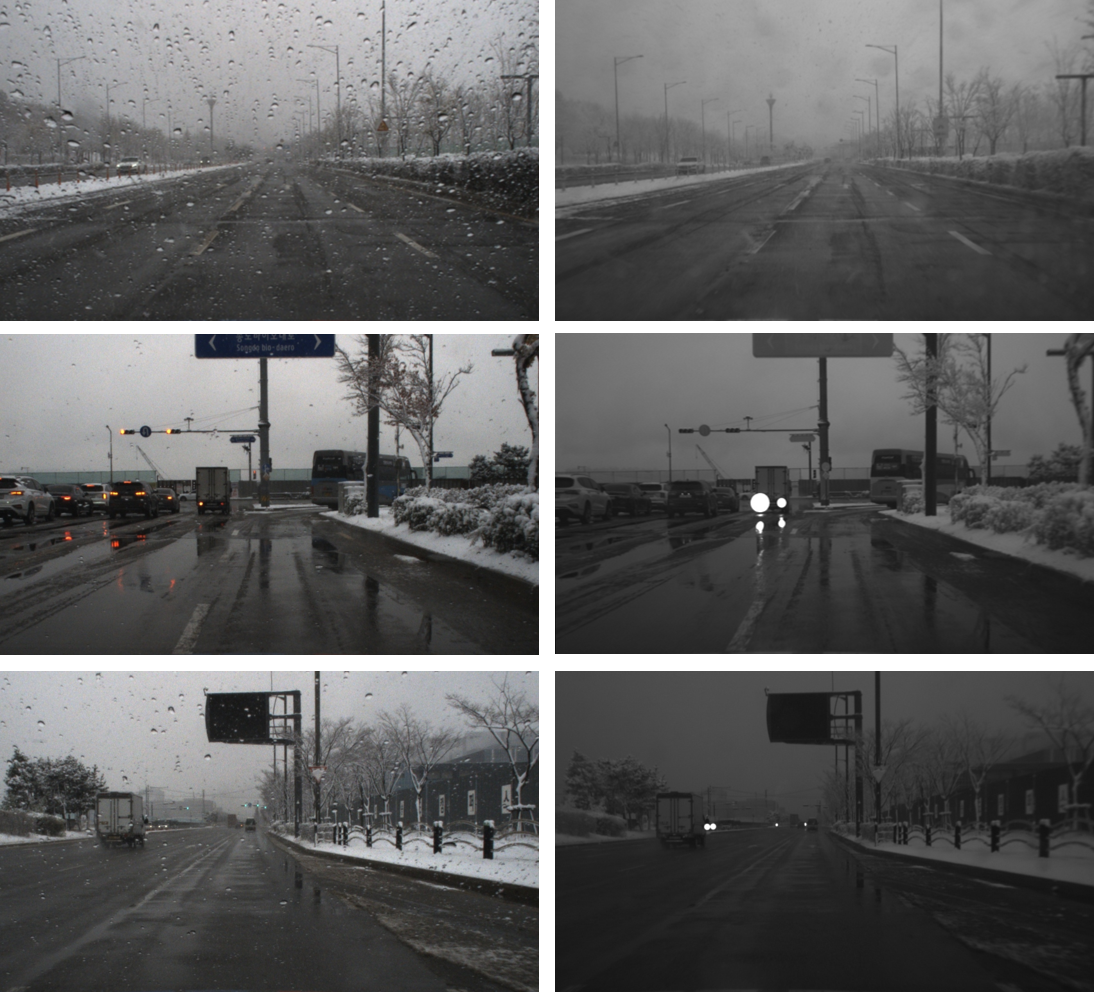}
        \subcaption{Driving Scene "Snowy"}
    \end{minipage}
    \caption{Overview of the RASMD dataset}
    \label{fig:dataset_ex}
\end{figure*}
To address the absence of publicly available SWIR datasets for autonomous driving research, we construct the RGB And SWIR Multispectral Driving (RASMD) dataset. This section provides a comprehensive overview of the data acquisition and calibration processes, annotation protocols, and the organization of the RASMD dataset for downstream tasks.

\subsection{Data Collection}

\begin{table}[ht]
    \renewcommand{\arraystretch}{2} 
    \resizebox{\columnwidth}{!}{%
    \begin{tabular}{l|lll}
        \hline
        \textbf{Sensor} & \textbf{Model} & \textbf{Frame Rate}  & \textbf{Characteristic} \\ \hline
        RGB Camera  & FLIR GS3-U3-32S4C-C  & max 120 FPS & 2048x1536 pixel      \\ \hline
        RGB Lens & \begin{tabular}[c]{@{}l@{}}EDMUND OPTICS 8.5mm C Series \\ Fixed Focal Length Lens\end{tabular} &  &   \\ \hline
        SWIR Camera & CREVIS HG-A130SW     & max 70 FPS  & 1296x1032 pixel    \\ \hline
        SWIR Lens   & COMPUTAR M0818-APVSW &             & 1000-1700nm long pass filter\\ \hline
    \end{tabular}%
}
    \caption{Specifications of the RGB and SWIR cameras used in our setup}
    \label{tab:camera_specs}
\end{table}

We created a data acquisition platform equipped with both RGB and SWIR sensors (\cref{tab:camera_specs}). Given the different frame rates of the cameras, precise time synchronization was a critical factor in the collection of synchronized views of the cameras. To manage this issue, we collected both images using a software trigger to ensure accurate synchronization between the two cameras. We collected 100K frames of multispectral driving data across diverse locations, lighting, and weather conditions. Specifically, we gathered synchronized multispectral data while driving through campus, city, and suburban areas to include diverse traffic situations. Additionally, we provide a range of weather variations like sunny, cloudy, rainy and snowy conditions. Table \ref{tab:data_extra} provides the distribution of data for each condition. With the RASMD dataset, we aim to assess and enhance the generalization and domain gap-handling abilities of deep learning networks for autonomous driving tasks. 


\begin{table}[h!]
\centering
\resizebox{\columnwidth}{!}{%
\begin{tabular}{|c|c|c|cc|cc|cc|}
\hline
Total acq. time &
  Total acq. distance &
  Total frame &
  \multicolumn{2}{c|}{Spectral range} &
  \multicolumn{2}{c|}{Location} &
  \multicolumn{2}{c|}{Weather condition} \\ \hline
\multirow{4}{*}{8.5 Hours} &
  \multirow{4}{*}{163.3 km} &
  \multirow{4}{*}{100K} &
  \multicolumn{1}{c|}{\multirow{2}{*}{RGB}} &
  \multirow{2}{*}{SWIR} &
  \multicolumn{1}{c|}{\multirow{2}{*}{Urban}} &
  \multirow{2}{*}{Suburban} &
  \multicolumn{1}{c|}{Sunny} &
  43.2k \\ \cline{8-9} 
 &
   &
   &
  \multicolumn{1}{c|}{} &
   &
  \multicolumn{1}{c|}{} &
   &
  \multicolumn{1}{c|}{Cloudy} &
  33.4k \\ \cline{4-9} 
 &
   &
   &
  \multicolumn{1}{c|}{\multirow{2}{*}{100k}} &
  \multirow{2}{*}{100k} &
  \multicolumn{1}{c|}{\multirow{2}{*}{56.2k}} &
  \multirow{2}{*}{43.8k} &
  \multicolumn{1}{c|}{Rainy} &
  10.7k \\ \cline{8-9} 
 &
   &
   &
  \multicolumn{1}{c|}{} &
   &
  \multicolumn{1}{c|}{} &
   &
  \multicolumn{1}{c|}{Snowy} &
  12.7k \\ \hline
\end{tabular}%
}
\caption{Data acquisition details and data distributions.}
\label{tab:data_extra}
\vspace{-0.6cm}   
\end{table}

\subsection{Image Alignment}
We collected images using two cameras with different optical parameters, distortions, and resolutions. To create a well-aligned dataset suitable for training, we needed to correct these differences and ensure pixel-wise alignment between the RGB and SWIR images. Our alignment process consisted of three key steps: calibration, feature-based alignment, and cropping.
\begin{figure*}[!ht]
    \centering
    \subfloat[Calibration]{%
        \includegraphics[trim={0 0.25cm 0 0.5cm},clip,height=0.19\textheight]{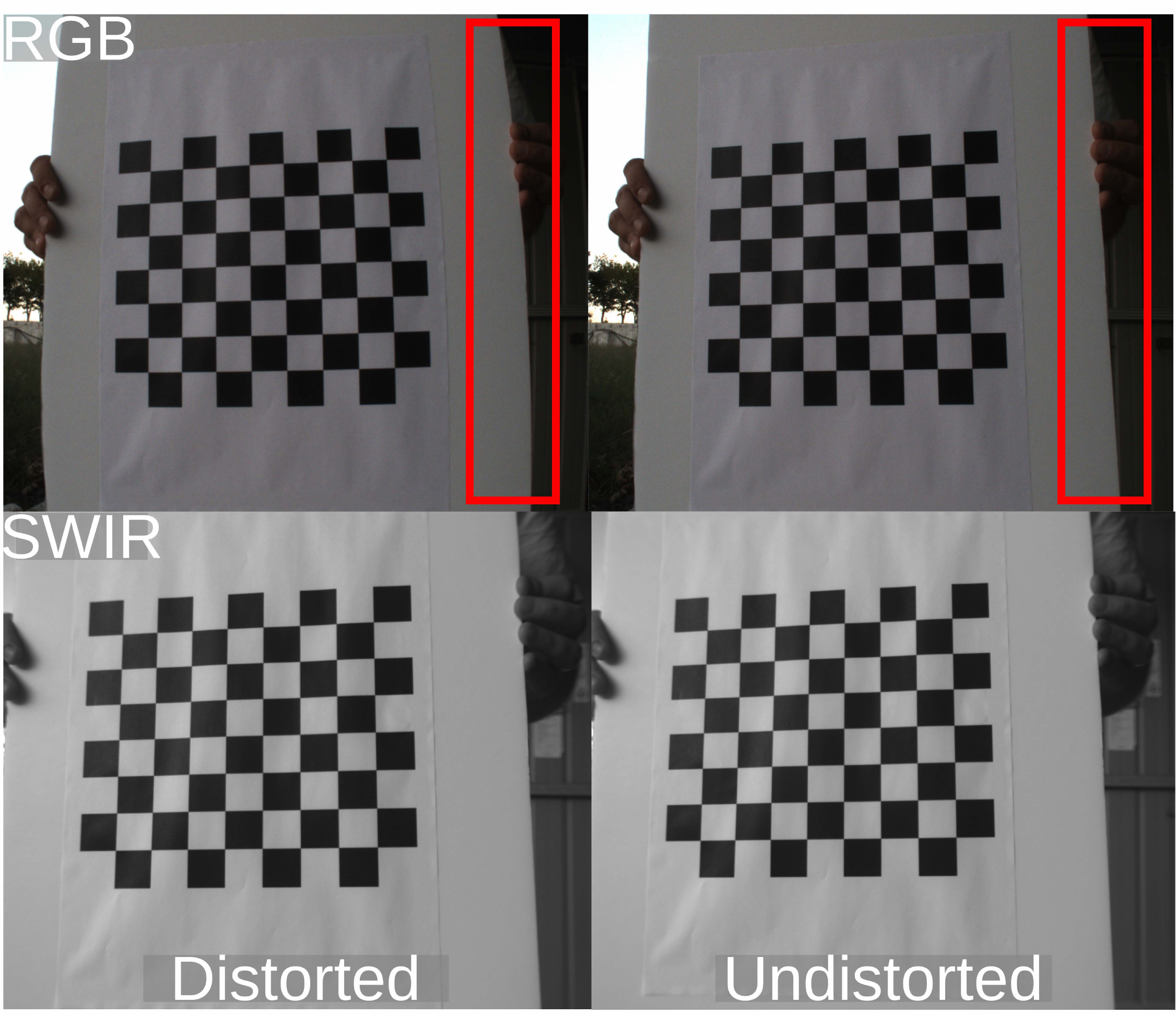}
        \label{fig:a_calibration}}
    \hspace*{\fill}
    \subfloat[Registration]{%
        \includegraphics[height=0.19\textheight]{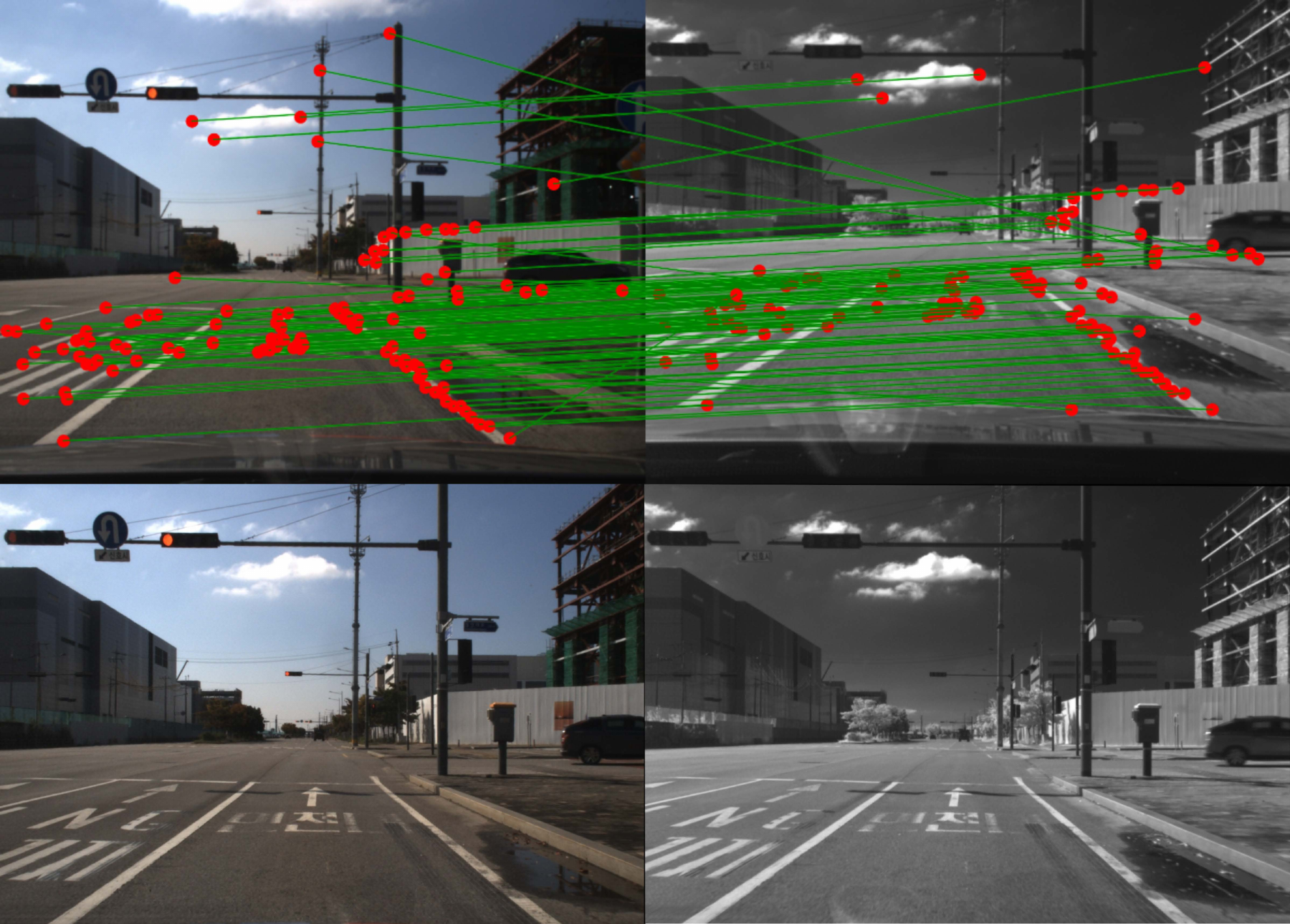}
        \label{fig:b_sift_matching}}
    \hspace*{\fill}
    \subfloat[Registered RGB and SWIR]{%
        \includegraphics[trim={0 2.2cm 0 0},clip,height=0.19\textheight]{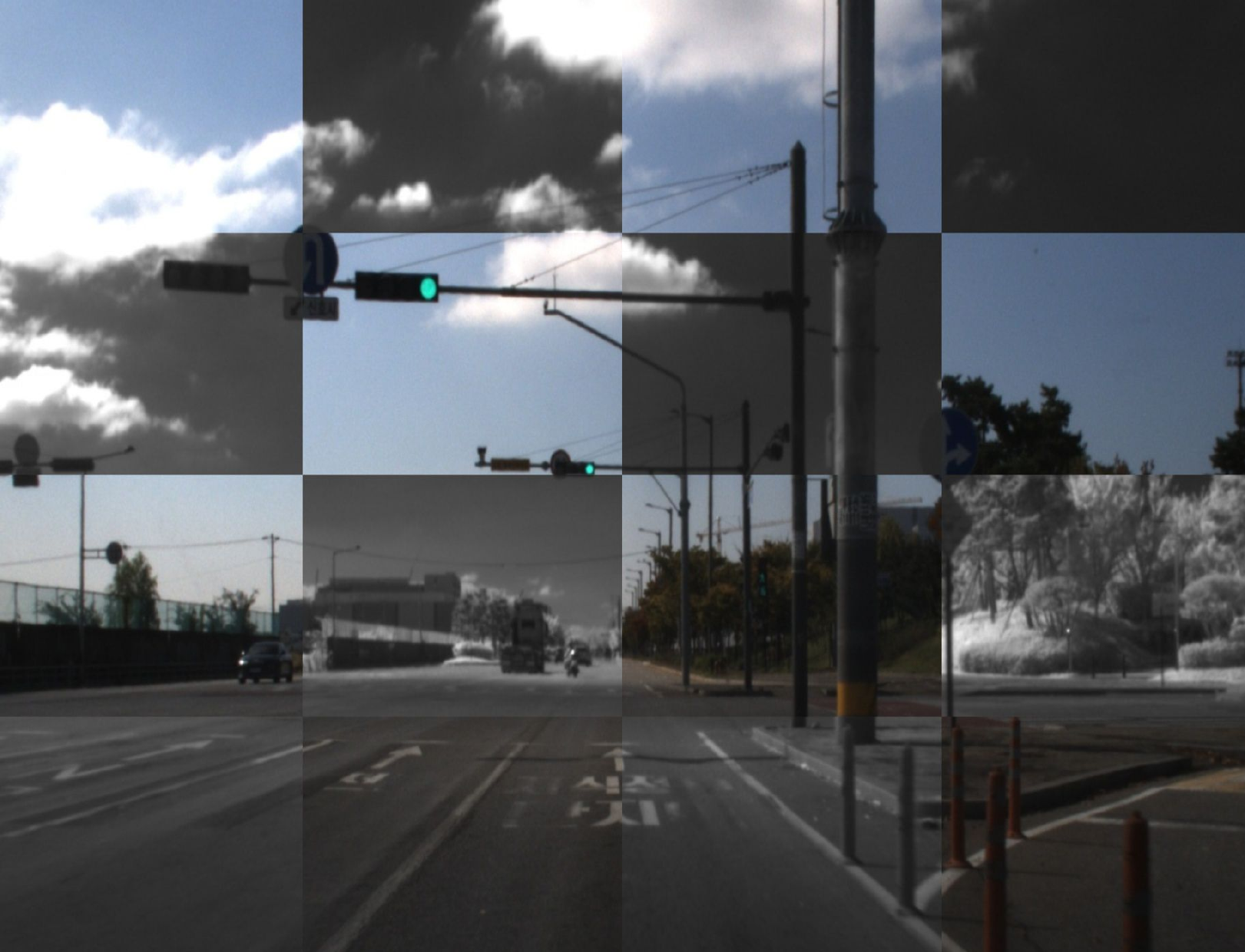}
        \label{fig:c_blended}}
    \caption{Camera calibration to correct lens distortions, demonstrated in image \textbf{(a)}, where the red bounding box highlights the corrected distortion in the RGB image. We employ SIFT feature matching for distortion correction, shown in \textbf{(b)}. In \textbf{(c)}, the visualization alternates between RGB and SWIR image patches in a checkerboard pattern, with each region representing the corresponding image part. The seamless transition at the center boundary indicates successful alignment between the two imaging modalities.}
    \label{fig:swir_rgb_registration}
\end{figure*}

\begin{figure*}[ht!]
     \centering
     \begin{subfigure}[b]{0.31\textwidth}
         \centering
        \includegraphics[width=\textwidth]{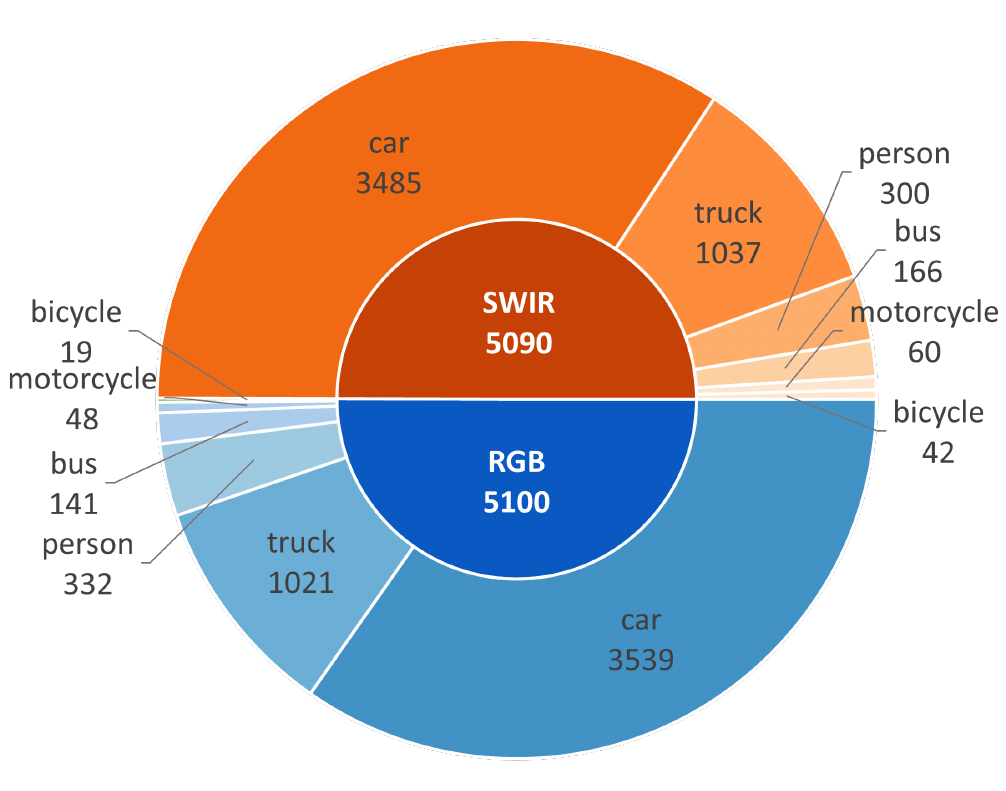}
         \caption{Train dataset}
         \label{fig:a_train_distribution}
     \end{subfigure}
     \hfill
     \begin{subfigure}[b]{0.31\textwidth}
         \centering
         \includegraphics[width=\textwidth]{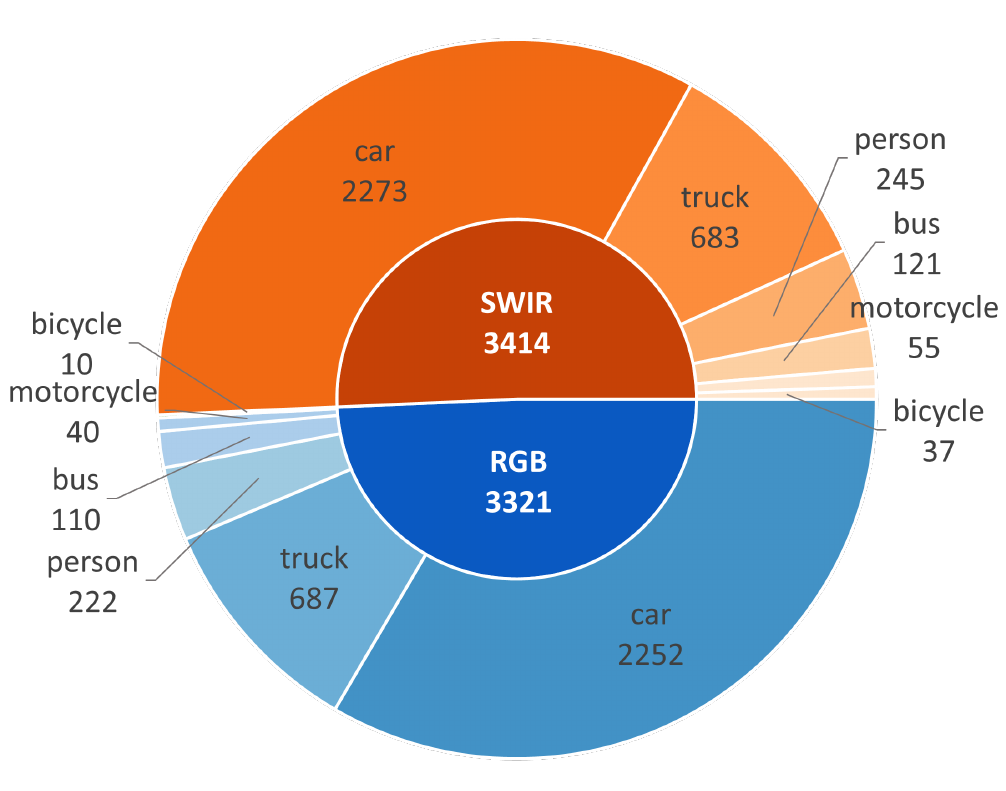}
         \caption{Test dataset}
         \label{fig:b_test_distribution}
     \end{subfigure}
     \hfill
     \begin{subfigure}[b]{0.31\textwidth}
         \centering
        \includegraphics[width=0.9\textwidth]{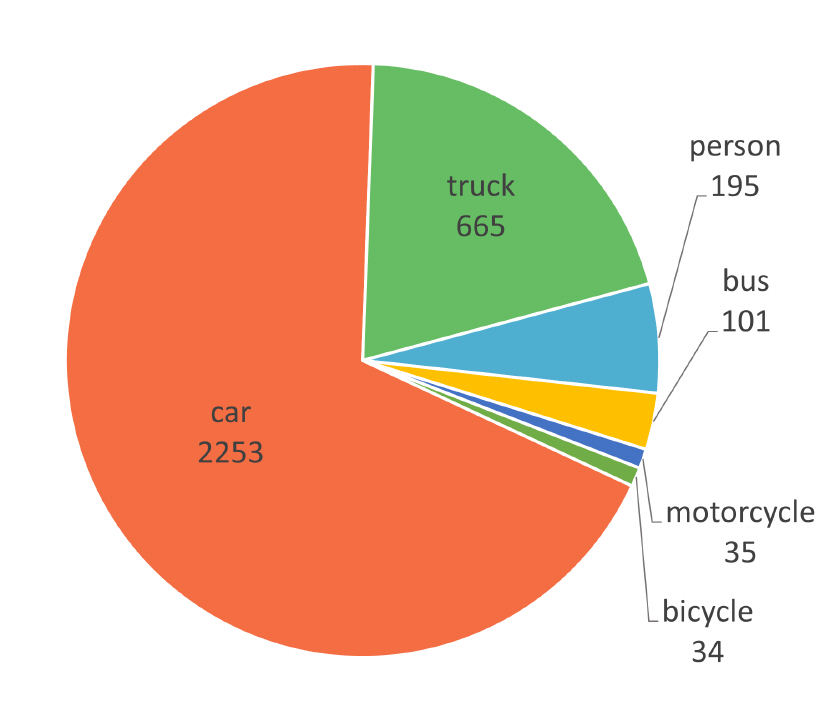}
         \caption{Merged labels test dataset}
         \label{fig:c_test_merged}
     \end{subfigure}
     \hfill
        \caption{RASMD dataset class distribution for detection labels: distribution of training \textbf{(a)} and testing \textbf{(b)} datasets with separate labels for SWIR (blue) and RGB (orange) images, and distribution of the merged labels dataset \textbf{(c)}. Differences in class counts arise from additional objects visible in the SWIR domain.}
        \label{fig:dataset_class_distribution}
\end{figure*}

To address intrinsic distortions unique to each camera, we performed geometric calibration using a 7 × 8 checkerboard pattern \cite{zhang_flexible_2000}. Given the SWIR camera’s wavelength sensitivity, a high-reflectance carbon-based ink-printed checkerboard was used, as conventional water-based ink patterns are not visible in the SWIR range. 
The undistorted images maintained the maximum field of view with minimal distortion, as shown in Fig. \ref{fig:a_calibration}.

While perfect alignment in non-planar scenes is challenging, our approach is effective for our specific imaging setup. The RGB and SWIR cameras were statically mounted with a fixed relative position, allowing us to compute a single homography matrix from a carefully selected image pair with strong feature correspondence. This homography transformation was applied uniformly across all images to ensure geometric consistency (see Fig. 4 and Supplementary Fig. 8).
For feature matching, we employed the Scale-Invariant Feature Transform (SIFT) algorithm \cite{lowe_sift_1999}, detecting key points across both RGB and SWIR images. Since feature repeatability can vary due to differences in wavelengths, we carefully selected image pairs with high feature correspondence to compute the homography transformation. Additionally, we applied RANdom SAmple Consensus (RANSAC) filtering to remove outlier matches and improve alignment robustness (Fig. \ref{fig:b_sift_matching}).
Once the homography transformation was applied, we cropped the images to the overlapping field of view, ensuring that both modalities shared a common, pixel-wise aligned region. This final step produced spatially registered image pairs suitable for multispectral analysis and machine learning applications (see Fig. \ref{fig:c_blended}). Since multi-modal image registration remains an active research area, we also provide unregistered image pairs for future studies. Further examples highlighting the robustness of the image alignment process are presented in Appendix.

\subsection{Annotations and Benchmark Tasks}\label{sec:annotation}

For the object detection task, we manually annotated a carefully selected subset of images that represent a range of challenging environmental conditions (e.g., low light, rain, fog, and backlighting). We focused on six common traffic object classes: car, truck, bus, bicycle, motorcycle, and person. To account for the unique visual characteristics of different imaging modalities, we performed separate annotations for the SWIR and RGB images. This produced two independent sets of training and testing data—one for each modality. In addition, to enable a fair cross-modality evaluation, we created a merged test dataset. We began with all object annotations from the RGB images and then supplemented these with additional annotations from the SWIR images that were not already present in the RGB dataset. In this way, the merged dataset additionally includes objects that are exclusively visible in the SWIR spectrum to provide a more comprehensive assessment of detection performance across both domains. 

Our dataset is organized as follows: the training and test set contains 1,432 and 956 images per modality, respectively, and the merged test set (for cross-domain evaluation) comprises 780 images. The distribution of object classes across these subsets is illustrated in Figure \ref{fig:dataset_class_distribution}.

 \section{Experiments}
\subsection{Object Detection}

To highlight the benefits of SWIR imaging in conditions where RGB detection often fails, we conducted object detection experiments on the RASMD dataset. These experiments focused on challenging scenarios such as fog, low lighting, and glare and demonstrated SWIR’s ability to detect crucial objects that RGB methods frequently miss under low-visibility conditions. 
\begin{table*}[!ht] 
    \centering
    \renewcommand{\arraystretch}{1.3}
    \resizebox{\textwidth}{!}{
    \Huge
        \begin{tabular}{c|*{7}{c}|*{7}{c}|*{8}{c}}
        \toprule 
                              \multirow{2}{*}{\centering \textbf{Method}} & \multicolumn{7}{c}{\textbf{SWIR domain}} 
                               & \multicolumn{7}{c}{\textbf{RGB domain}} 
                               & \multicolumn{8}{c}{\textbf{Ensemble}} \\
                               & $\mathrm{AP}_\mathrm{person}$ & $\mathrm{AP}_\mathrm{car}$ & $\mathrm{AP}_\mathrm{truck}$  & $\mathrm{AP}_\mathrm{bus}$ & $\mathrm{AP}_\mathrm{bicycle}$  & $\mathrm{AP}_\mathrm{m.cycle}$ & $\mathrm{mAP}$ 
                               & $\mathrm{AP}_\mathrm{person}$ & $\mathrm{AP}_\mathrm{car}$ & $\mathrm{AP}_\mathrm{truck}$  & $\mathrm{AP}_\mathrm{bus}$ & $\mathrm{AP}_\mathrm{bicycle}$  & $\mathrm{AP}_\mathrm{m.cycle}$ & $\mathrm{mAP}$ 
                               & $\mathrm{AP}_\mathrm{person}$ & $\mathrm{AP}_\mathrm{car}$ & $\mathrm{AP}_\mathrm{truck}$  & $\mathrm{AP}_\mathrm{bus}$ & $\mathrm{AP}_\mathrm{bicycle}$  & $\mathrm{AP}_\mathrm{m.cycle}$ & $\mathrm{mAP}$ &$\Delta \mathrm{mAP}$    \\
        \midrule
        Faster-RCNN\cite{faster_rcnn}  & 0.4017 & 0.4003 & 0.6391 &  0.3765 & 0.2644 & 0.6391 & 0.3721  & 0.4628 & 0.5841 & 0.7869 & 0.5501&  0.0564& 0.3204 & 0.4601& 0.5079 & 0.6074 &  0.8024 &  0.5948 & 0.3068 & 0.2844  & 0.5173 & \textcolor{green}{$\uparrow$+0.0572} \\
        SSD\cite{ssd}  & 0.2750 &0.3762 & 0.5947 & 0.3828 & 0.1053& 0.1312& 0.3109 &0.3333 & 0.5621 & 0.7575 & 0.5317 & 0.0594 & 0.2831 & 0.4212& 0.3558 & 0.5637 & 0.7423 & 0.5512 & 0.1341 & 0.2451 & 0.4320 & \textcolor{green}{$\uparrow$+0.0108} \\
        Centernet\cite{centernet} & 0.3619 & 0.3815 &  0.5540 & 0.3211 & 0.2796 & 0.2626& 0.3601 & 0.3789 & 0.5537 & 0.7124 & 0.5643 & 0.0594 & 0.2891 & 0.4263 & 0.4302 & 0.5598 &  0.7325 & 0.5318 & 0.3279 & 0.4537 & 0.5060 & \textcolor{green}{$\uparrow$+0.0797} \\
        DETR\cite{detr}    & 0.3808 & 0.3589 & 0.6147 &  0.3839 & 0.2735 &  0.2356 & 0.3746& 0.4839 & 0.6180 &  0.8070 &  0.5964 & 0.1147 &  0.3562 & 0.4960& 0.4947 &  0.6046 & 0.8163 & 0.6223 & 0.3155 &  0.3898 & 0.5405 & \textcolor{green}{$\uparrow$+0.0445} \\
        Deformable DETR\cite{deformable_detr}  & 0.3275 & 0.3329 &  0.5335 & 0.2012 &  0.1853 & 0.1973& 0.2963& 0.3717 & 0.6119 & 0.7627 & 0.5594 & 0.0772 & 0.3094 & 0.4487 & 0.3888 & 0.5746 & 0.7219& 0.5474 & 0.2418 & 0.3595 & 0.4723 & \textcolor{green}{$\uparrow$+0.0236} \\
        Conditional DETR\cite{conditional_detr}  &  0.3657 & 0.4192 & 0.6374 &  0.4181 & 0.1448 & 0.2145 & 0.3666& 0.3788 &  0.5940&  0.7479&0.5455  &0.1386  &0.4079  & 0.4688 &  0.4297  &  0.6036 & 0.7619 & 0.5643 & 0.2115  & 0.4799  & 0.5085 & \textcolor{green}{$\uparrow$+0.0397} \\
        YOLOv7\cite{yolov7}  & 0.3636 & 0.4092 & 0.6432 & 0.3975 & 0.2405 & 0.3209 & \textbf{0.3958}& 0.4554 & 0.5771 & 0.7625 & 0.5687 & 0.1570 & 0.4004 & 0.4869& 0.4744 & 0.5887 & 0.7510 & 0.5970 & 0.3199 & 0.4580 & 0.5315 & \textcolor{green}{$\uparrow$+0.0446} \\
        DINO\cite{dino}    &  0.3252 & 0.3863  & 0.6440 & 0.3960 & 0.3345 & 0.2814 & 0.3945 & 0.4875 & 0.6334 & 0.8086 & 0.6139 & 0.1015 & 0.3749 & \textbf{0.5033}& 0.5250 & 0.6220 & 0.7919 & 0.5467 & 0.4089 & 0.4581 & \textbf{0.5588} & \textcolor{green}{$\uparrow$+0.0555}\\
        Co-DETR\cite{codetr} & 0.2945 & 0.3210 & 0.5293& 0.2671&  0.0653&  0.0165&  0.2490& 0.4837 & 0.6053 & 0.7083 & 0.4423 & 0.0297 & 0.3204 & 0.4316 & 0.5051 & 0.6030 & 0.7025 & 0.4540 & 0.0990 & 0.2270 & 0.4318 & \textcolor{green}{$\uparrow$+0.0002} \\
        \bottomrule
    \end{tabular}
    }
    \caption{Object detection results of widely used models in the literature. Ensembling RGB with SWIR yields superior performance compared to using RGB alone. The green text highlights improvements over the RGB-only results.}
    \label{tab:detection_all}
\end{table*}

\begin{figure*}[ht!]
    \centering
    \includegraphics[width=1\linewidth]{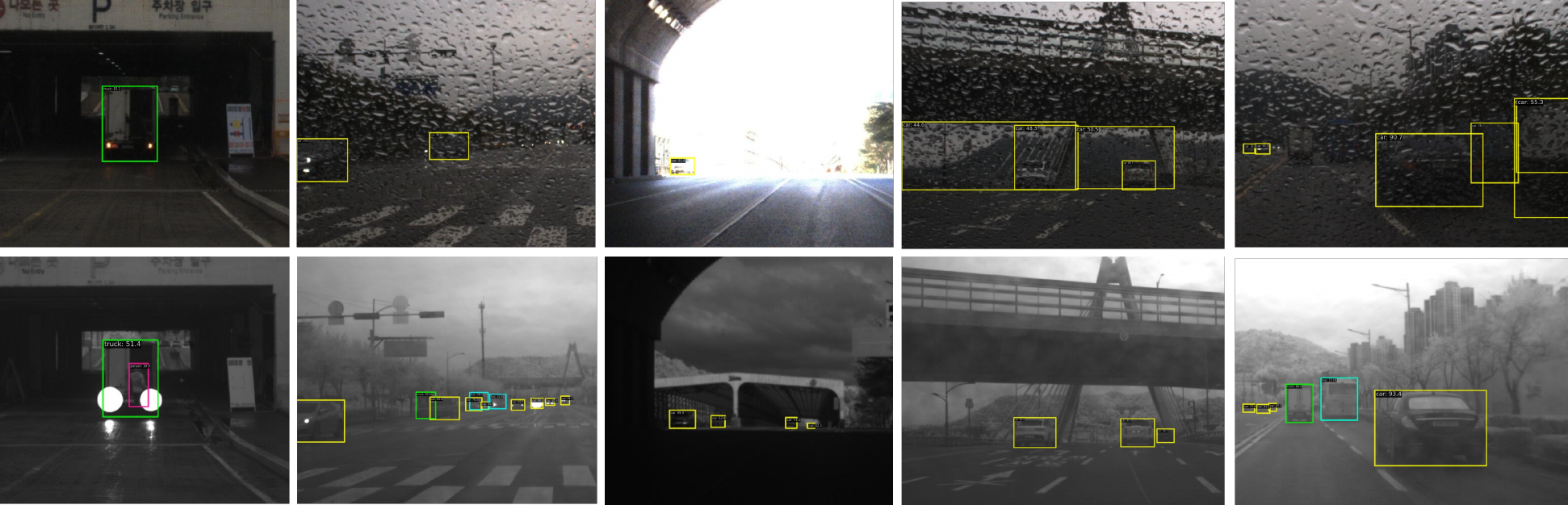}
    \caption{Examples of detection results of the RASMD object detection subset, evaluated on separate test data for RGB (first row) and SWIR (second row) images. The advantages of SWIR imaging under challenging conditions are clearly visible in comparison with the RGB images.}
    \label{fig:detection_result}
\end{figure*}

For each condition, we trained separate detection models using RGB and SWIR data. Their outputs were then combined using an ensemble approach with non-maximum suppression at an IoU threshold of 0.5. By leveraging SWIR’s robustness, this method compensates for the performance decline often observed in RGB detection under challenging conditions, effectively harnessing the strengths of both spectral bands. As shown in the ensemble section of Table \ref{tab:detection_all}, combining RGB and SWIR detection outputs significantly enhances performance compared to using RGB alone. This improvement is particularly notable for vulnerable road users (VRU), such as pedestrians and cyclists, where substantial performance gains are observed. The higher mean Average Precision (mAP) scores across multiple object categories further demonstrate that this fusion effectively compensates for scenarios in which RGB detection underperforms. Additional detection results are provided in Appendix.

\subsection{RGB to SWIR Translation}

\begin{figure*}[h!]
    \centering
    \includegraphics[width=1\linewidth]{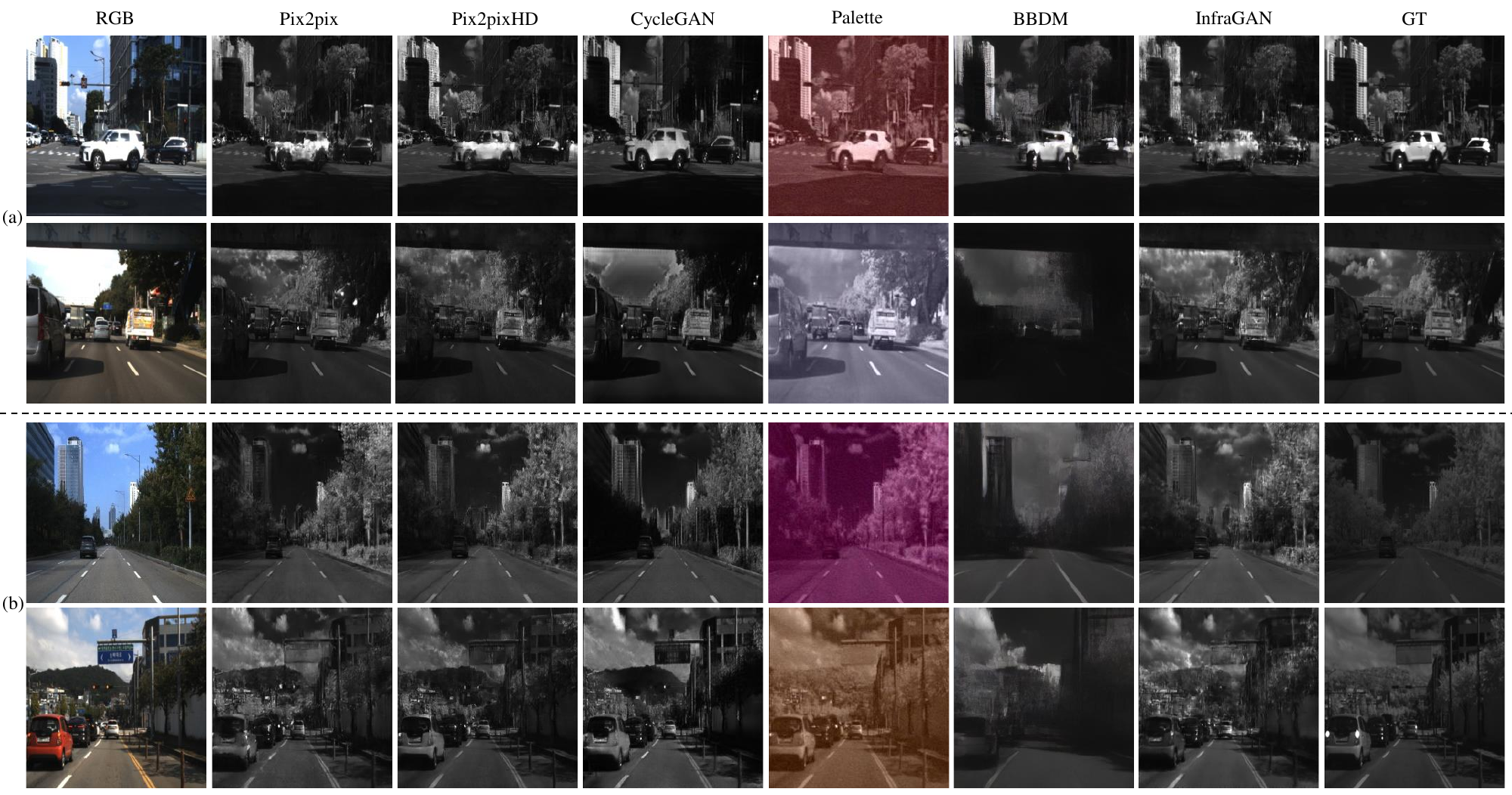}
    \caption{Examples of RGB-to-SWIR translation results on the RASMD dataset. \textbf{(a)} presents the results of models trained from scratch, and \textbf{(b)} shows the zero-shot translation results of the same models.  Both BBDM and Pix2pixHD demonstrate comparable performance in generating SWIR images and capturing critical details effectively. In contrast, Palette, not designed for multispectral image generation, struggles to produce accurate SWIR representations.}
    \label{fig:translation_vis}
\end{figure*}

Since data availability is really important for training robust deep learning models, some studies tried to overcome the lack of data in the infrared (IR) spectrum by approaches such as knowledge distillation \cite{d3t,thermal_sam} and domain translation \cite{rgb2nir, c2sal, vq-infratrans, infragan, thermalgan}. Among them, research on translating RGB images to IR domains has gained attention to enable the scaling of datasets without the need for time-consuming and costly data acquisition and annotation processes. 

However, as we mentioned in previous sections, due to the lack of data on the SWIR spectrum, no other study evaluated their methods on the SWIR range. To this aspect, we created a subset from RASMD for the RGB-to-SWIR image translation task, comprising 3,900 images for training, 979 for testing, and 930 for zero-shot testing. We utilized this dataset to evaluate existing I2I translation methods in the literature. All the comparison experiments are performed on the default parameters of the methods. Visual comparisons of the translated images are presented in \cref{fig:translation_vis}, while quantitative comparisons are detailed in \cref{tab:i2i_comparison} and \cref{tab:i2i_comparison_zeroshot}. These results demonstrate that our dataset is well-aligned and highlight its potential as a benchmark for future RGB-to-SWIR domain translation techniques. Additional examples of translated images can be found in Appendix.

\begin{table}[ht]
\centering
\renewcommand{\arraystretch}{1.5}
\resizebox{\columnwidth}{!}{%
\begin{tabular}{lccccccc}
\hline
Method    & Type & PSNR↑ & SSIM↑  & RMSE↓ & FID↓  & LPIPS↓ & DISTS↓ \\ \hline
Pix2pix \cite{pix2pix}     & G & 28.48 & 0.8514 & 5.04 & 32.92 & 0.0847 & 0.12  \\
Pix2pixHD \cite{pix2pixhd} & G & 30.50 & 0.8897 & 4.79 & 35.35 & 0.0635 & 0.1289 \\
CycleGAN \cite{cyclegan}   & G & 20.34 & 0.6078 & 8.28 & 62.74 & 0.2078 & 0.193 \\
BBDM \cite{bbdm}           & D & 31.06 & 0.8824 & 4.55 & 28.88 & 0.0763 & 0.1133 \\
Palette \cite{palette}     & D & 12.84 & 0.5221 & 9.81 & 112.45& 0.3619 & 0.2898 \\
InfraGAN \cite{infragan}   & G & 29.08 & 0.8654 & 5.24 & 31.04 & 0.0746 & 0.1206 \\
\hline 
\end{tabular}%
}
\caption*{\footnotesize \tiny \raggedright Type G: GAN based method, Type D: Diffusion based method.}
\caption{RGB to SWIR translation performance comparison with various I2I translation methods on Our RASMD dataset.}
\label{tab:i2i_comparison}
\end{table}

\begin{table}[ht]
\centering
\renewcommand{\arraystretch}{1.5}
\resizebox{\columnwidth}{!}{%
\begin{tabular}{lccccccc}
\hline
Method    & Type & PSNR↑ & SSIM↑  & RMSE↓ & FID↓  & LPIPS↓ & DISTS↓ \\ \hline
Pix2pix \cite{pix2pix}     & G & 18.46 & 0.5510 & 9.40 & 61.16 & 0.2255 & 0.2058  \\
Pix2pixHD \cite{pix2pixhd} & G & 19.44 & 0.5883 & 9.23 & 68.45 & 0.2251 & 0.2180  \\
CycleGAN \cite{cyclegan}   & G & 16.14 & 0.4786 & 10.06& 44.16 & 0.2450 & 0.2109  \\
BBDM \cite{bbdm}           & D & 17.23 & 0.5162 & 9.77 & 147.57& 0.3859 & 0.3077 \\
Palette \cite{palette}     & D & 11.20 & 0.4348 & 10.04& 114.15& 0.4470 & 0.3093 \\
InfraGAN \cite{infragan}   & G & 17.84 & 0.5281 & 9.62 & 62.81 & 0.2259 & 0.2162 \\ 
\hline
\end{tabular}%
}
\caption*{\footnotesize \tiny \raggedright Type G: GAN based method, Type D: Diffusion based method.}
\caption{RGB to SWIR Zeroshot translation performance comparison with various I2I translation methods on Our unseen data.}
\label{tab:i2i_comparison_zeroshot}
\end{table}

\section{Conclusion}
In this paper, we introduced the RGB and SWIR Multispectral Driving (RASMD) dataset, which was created to address the limited availability of SWIR data for driving scenes. RASMD aims to enable in-depth analysis and practical applications of SWIR wavelength characteristics to support research beyond conventional RGB imaging to achieve robust performance in challenging conditions. Our object detection experiments on this dataset confirmed the advantages of SWIR imaging in scenarios where RGB camera may face limitations. Additionally, our experiments with various image translation methods highlight the potential to generate SWIR images from RGB, offering a promising avenue for data scale-up. We anticipate that RASMD will foster research on multispectral imaging for autonomous systems, particularly in complex driving environments utilizing SWIR imaging, despite the high cost of SWIR sensors.

In future work, we plan to expand the dataset by increasing the number of annotated images and broadening the range of object classes to include additional elements critical for autonomous driving, such as traffic signs, traffic lights, and road markings. We aim to scale up the dataset by incorporating additional weather conditions and environmental scenarios. Specifically, we will add weather severity labels to images and plan to incorporate semantic segmentation annotations to create a comprehensive benchmark for SWIR imaging.

{
    \small
    \bibliographystyle{ieeenat_fullname}
    \bibliography{main}
}


\end{document}